\documentclass[sigconf]{acmart}
\usepackage[utf8]{inputenc}
\usepackage{textgreek}
\usepackage{booktabs}
\usepackage{xcolor}
\usepackage[normalem]{ulem}
\usepackage{colortbl}
\usepackage{multirow}
\usepackage{array}
\usepackage{makecell}
\usepackage{tabularx}
\usepackage[most]{tcolorbox}
\usepackage{circledsteps}
\usepackage{graphicx}
\usepackage{subcaption}
\usepackage{float}
\usepackage{wrapfig}
\usepackage{algorithm}
\usepackage{algorithmic}
\usepackage[dvipsnames]{xcolor}
\usepackage{pifont}
\usepackage{makecell}
\usepackage{booktabs,longtable,array}

\usepackage{balance}

\PassOptionsToPackage{dvipsnames,table,xcdraw}{xcolor}
\usepackage[normalem]{ulem}
\definecolor{myblue}{RGB}{0,0,0}
\AtBeginDocument{%
  }
    
\definecolor{lightgray}{gray}{0.95} 
\newtcolorbox{rqbox}{
  colback=gray!20,       
  colframe=black,     
  boxrule=0.5pt,        
  arc=2pt,              
  left=6pt, right=6pt,  
  top=4pt, bottom=4pt,  
}

\copyrightyear{2026}
\acmYear{2026}
\setcopyright{cc}
\setcctype{by}
\acmConference[UbiComp Companion '26]{Companion of the 2026 ACM International Joint Conference on Pervasive and Ubiquitous Computing}{October 11--15, 2026}{Shanghai, China}
\acmBooktitle{Companion of the 2026 ACM International Joint Conference on Pervasive and Ubiquitous Computing (UbiComp Companion '26), October 11--15, 2026, Shanghai, China}
\acmDOI{10.1145/3798063.3839206}
\acmISBN{979-8-4007-2533-3/2026/10}

\copyrightyear{2026}
\acmYear{2026}
\setcopyright{cc}
\setcctype{by}
\acmConference[UbiComp Companion '26]{Companion of the 2026 ACM International Joint Conference on Pervasive and Ubiquitous Computing}{October 11--15, 2026}{Shanghai, China}
\acmBooktitle{Companion of the 2026 ACM International Joint Conference on Pervasive and Ubiquitous Computing (UbiComp Companion '26), October 11--15, 2026, Shanghai, China}
\acmDOI{10.1145/3798063.3839206}
\acmISBN{979-8-4007-2533-3/2026/10}

\begin{document}

\title[How Uncertainty Estimation Improves Proactive Assistant Alignment]{Vision: "What Don't I Know About User's Preferences?" How Uncertainty Estimation Improves Proactive Assistant Alignment}



\author{Zhiyu An}
\affiliation{\institution{University of California, Merced}
\city{Merced}
\state{California}
\country{USA}}

\author{Ziyi Xuan}
\orcid{0000-0003-1375-3645}
\affiliation{\institution{Lehigh University}
\city{Bethlehem}
\state{Pennsylvania}
\country{USA}
}

\author{Yu Yang} 
\orcid{0000-0003-1627-5503}
\affiliation{\institution{Lehigh University}
\city{Bethlehem}
\state{Pennsylvania}
\country{USA}
}

\author{Wan Du}
\affiliation{\institution{University of California, Merced}
\city{Merced}
\state{California}
\country{USA}}

\settopmatter{authorsperrow=4}

\begin{abstract}
Proactive Smart Assistants (PSAs) increasingly use general knowledge, learned patterns from group-level data, patterns from individual user's data, and recent sensor observations to anticipate which action a user may prefer. 
Existing systems typically select actions from predicted user preference likelihoods, but these likelihoods alone do not reveal how reliable the prediction is, which evidence supports it, or why ambiguity remains. 
In this vision paper, we argue that uncertainty estimation can provide this missing information and substantially change appropriate downstream behavior. 
Using the example of a home assistant, we introduce the questions that uncertainty estimation aims to answer, i.e. how sure is the assistant about the predicted preference likelihood, whether the confidence come from group-level assumptions or individual's data, and whether likelihood ambiguity reflect the user's intrinsic preference variability or insufficient data for the particular situation.
We further introduce uncertainty estimation applications in neighboring domains, such as human trajectory prediction, personalized activity recognition, information-collection, and model-based control.
Finally, we draw on insights from neighboring domains to envision how future uncertainty-aware proactive assistants may address key challenges in proactive interaction.
\end{abstract}



\maketitle

\section{Introduction}

\begin{figure*}
    \centering
    \includegraphics[width=\textwidth]{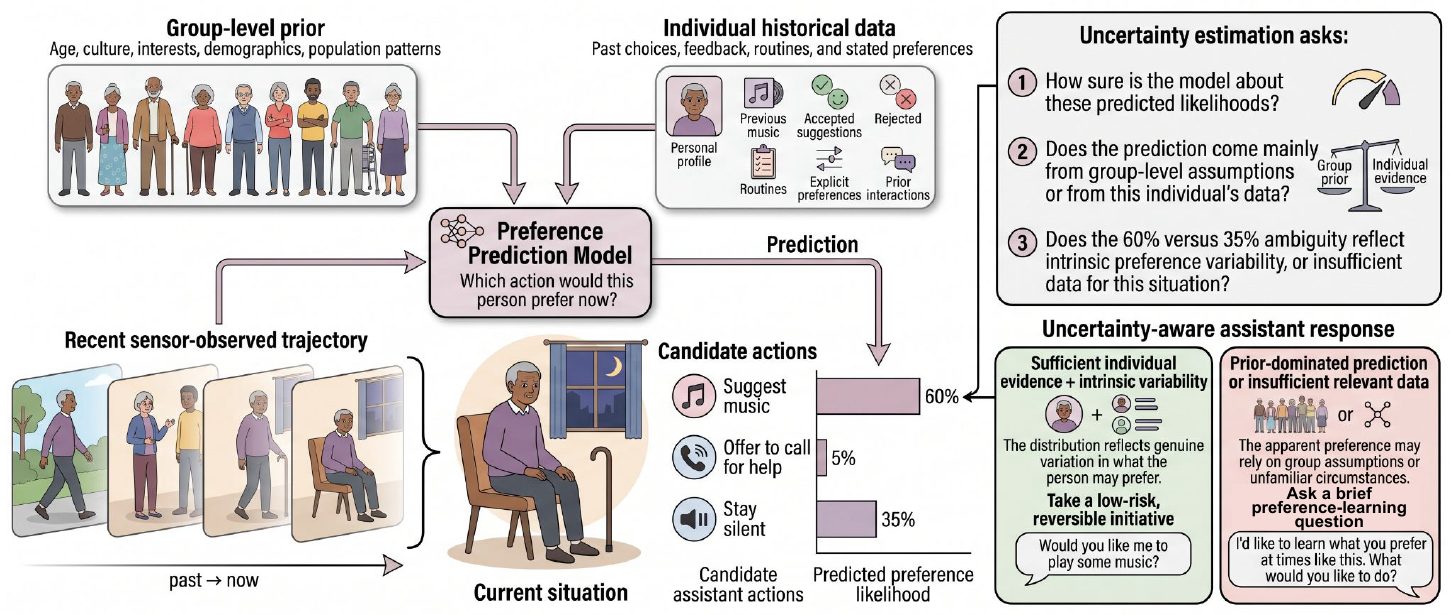}
    \caption{Uncertainty estimation can substantially inform proactive assistant and change its proper downstream actions.}
    \label{fig:teaser example}
\end{figure*}

Research on Proactive Smart Assistants (PSAs) has examined both domain-specific \cite{janssen2020virtual} and general-purpose assistants \cite{triguero2024general}, assistants personalized for a single user \cite{xuan2026after, xuan2025design} and assistants designed to mediate and facilitate collaboration of multiple users \cite{liu2024xiang, wang2026seesawbot}, and for embodies ranging from a smart speaker \cite{cha2020hello, deng2024towards} to smart glasses and phones \cite{yang2025proagent}. 
However, despite Human-AI interaction has been traditionally formulated as decision-making under uncertainty \cite{horvitz1999principles, amershi2019guidelines}, there is limited progress in incorporating uncertainty estimation in the design of PSA.

\textbf{Existing work} such as
\textit{Better To Ask Than Assume} \cite{oh2024better} ask the user for preference, but the goal was to preserve user agency rather than improving PSA action's alignment with the user.
\textit{Navigating User-System Gaps} \cite{kim2024navigating} aims to improve PSA alignment, but focused primarily on investigating how user perceive and cope with the gap.
\textit{PACT} \cite{he2026pact} makes an explicit ask-or-act decision based on whether the available context is sufficient, but its uncertainty concerns human intent and task requirements, rather than calibrated personalized preferences over actions.
\textit{GLOBE} \cite{gonzalez2026personalized} uses confidence to determine whether to invoke an LLM for additional reasoning, but it uses uncertainty to control computational routing, not whether the assistant should intervene, ask the user, or remain silent.
\textit{SLaTe-PRO} \cite{patel2023predicting} recognizes that some human behavior is inherently stochastic, but does not distinguish between inherent stochasticity with lack-of-data.
Finally, \textit{FAMER} \cite{wang2025strangers} models unknown, implicit user desires and uses communication to gather information, but it uses LLM reflection and does not model different source of evidence and the uncertainty of predictions.

Incorporating uncertainty estimation in the design of PSAs has the potential to substantially change its proper downstream actions and improve User-PSA alignment.
We illustrate a concrete example in Figure \ref{fig:teaser example}.
In this example, a general-purpose home assistant considers a set of candidate proactive actions, and uses group-level data prior, individual historical data, and recent sensor-observed trajectory to predict the likelihood of the user preferring each candidate action.
Instead of interpreting the likelihood distribution, \textbf{uncertainty estimation considers the uncertainty of the likelihoods themselves}. 
Uncertainty estimation aims to answer i) How sure is the model about the predicted likelihoods? ii) Does the prediction come mainly from group data priors and in-group consistency assumptions, or from this particular individual's observed evidences? iii) Does the ambiguity in the likelihood distribution reflect observed preference variability, or does it reflect insufficient data for this particular situation?
Different answers to these implications leads to substantially different proper downstream actions.
If the assistant has sufficient evidence of a observed, intrinsic variability in this particular user's preference, the proper action might be to directly initiate a reversible, high-likelihood action.
If, instead, the likelihoods are mainly "guessed" from group data priors or from a lack of data, then the proper downstream action might be to ask a preference-eliciting question aimed at learning user's preference at this particular situation.

In this paper, we discuss the use case of uncertainty estimation in domains similar to PSA, and draw connections from these domains to potential use cases within PSA.

\section{Applied Uncertainty Estimation In Neighboring Domains}

Several neighboring domains face a problem similar to that of proactive smart assistants: predicting human behavior from noisy, partial, and context-dependent observations. 
Research on pedestrian and vehicle trajectory forecasting has particularly emphasized the multimodality of human behavior.
\textit{Social GAN} \cite{gupta2018social} generates multiple socially plausible future trajectories instead of committing to a single deterministic prediction.
\textit{Trajectron++} \cite{salzmann2020trajectron++} further demonstrates that probabilistic human-behavior prediction can integrate heterogeneous evidence, including past trajectories, agent interactions, semantic maps, and the planned behavior of the observing system.
Driving behavior prediction methods such as \textit{MultiPath} \cite{chai2019multipath} separates uncertainty over discrete behavioral modes from uncertainty within each mode by predicting probabilities over anchor trajectories together with a continuous distribution around each anchor.
Beyond uncertainty associated with an individual, \textit{Tang et al.} \cite{tang2021collaborative} introduce collaborative uncertainty to capture additional uncertainty induced by interactions and correlations among multiple agents.
These works demonstrate the importance of predicting a distribution over possible human behaviors; however, such a distribution does not necessarily indicate how reliably its probabilities have been estimated.

Other work explicitly models different sources of uncertainty.
Most closely related to our proposed evidence-source analysis, \textit{Itkina and Kochenderfer} \cite{itkina2023interpretable} attribute trajectory-prediction uncertainty to interpretable input concepts, including past agent behavior, road structure, and social context.
\textit{An et al.} \cite{an2025disentangling} propose the Compressed Data Representation Model to disentangle aleatoric uncertainty arising from inherent system randomness from epistemic uncertainty arising from insufficient data.
Together, these works suggest that a PSA should not report only an aggregate confidence value: it should identify both which evidence sources support a prediction and whether the remaining uncertainty reflects genuine variability or insufficient knowledge about the individual and current context.

Human-motion prediction and wearable activity recognition further distinguish uncertainty caused by limited knowledge from variability inherent in observed human behavior.
\textit{Xu et al.} \cite{xu2021probabilistic} explicitly separate epistemic uncertainty, which indicates unfamiliar observations or insufficient model knowledge, from heteroscedastic aleatoric uncertainty associated with variation in possible human motions.
In wearable activity recognition, \textit{Akbari and Jafari} \cite{akbari2020personalizing} use epistemic and aleatoric uncertainty to identify informative examples for personalizing a population-trained model to a new user.
\textit{Huo et al.} \cite{huo2020uncertainty} use uncertainty quantification to identify unknown contexts in mobile activity recognition, recognizing that behavior observed in one situation may not generalize reliably to a new context.
These results are relevant to PSAs initialized from group-level data: uncertainty can indicate that a prediction is supported primarily by a population model and that additional individual or situation-specific evidence is needed.

Uncertainty estimation can influence downstream information gathering and action selection.
Uncertainty can directly constrain downstream actions: \textit{CLUE} \cite{an2023clue, ding2025safe} incorporates epistemic uncertainty into model-predictive control by filtering highly uncertain action trajectories, favoring actions with both high utility and confidence, and reverting to a reliable fallback controller when no sufficiently safe action is available.
\textit{Ardywibowo et al.} \cite{ardywibowo2019adaptive} use activity-prediction entropy to adaptively select sensor inputs while balancing uncertainty reduction against sensing costs.
\textit{TIME} \cite{hou2026time} similarly uses periods in which the system is not mission-critical to select exploratory actions that balance expected information gain against cost.
\textit{Dixit et al.} \cite{dixit2023adaptive} demonstrate how adaptively calibrated uncertainty sets for human or agent trajectories can be passed directly to a downstream planner, allowing system behavior to respond to changes in prediction reliability.
Analogously, a PSA could use the source and type of its preference uncertainty to decide whether to initiate a reversible action, continue observing, elicit the user's preference, or remain silent.

Finally, \textit{Learning Progress Monitoring} \cite{hou2025beyond} shows that exploration based on raw prediction error can become fixated on irreducible randomness; instead, its method rewards improvements in the learned model, prioritizing observations from which the agent can still learn.
Following that intuition, a PSA should gather additional evidence when its uncertainty is reducible, but should not repeatedly query or stay silent and observe the user when the remaining ambiguity reflects genuine preference variability.

\section{Envisioning Future Uncertainty-Aware Proactive Assistants}

The neighboring domains reviewed above suggest a broader vision for how uncertainty could shape future proactive assistants. Rather than treating uncertainty as an additional confidence value attached to a predicted action, future PSAs may use it to decide how to participate in an interaction: whether to act, suggest, ask, wait, or remain silent. This vision extends longstanding ideas in mixed-initiative interaction, in which intelligent systems consider the costs of acting, interrupting, and requesting clarification \cite{horvitz1999principles}, as well as human--AI interaction guidelines that encourage systems to limit automation when uncertain and support efficient correction \cite{amershi2019guidelines}. Uncertainty estimation could make these principles more actionable by revealing not only that an assistant is uncertain, but also what remains unknown and whether further interaction is likely to resolve it.

\textbf{Initiative that adapts to the source of uncertainty.}
Work on selective prediction and uncertainty-aware control often frames uncertainty as a choice between acting and deferring \cite{an2023clue,dixit2023adaptive,bondi2022role}. Future PSAs could support a richer range of responses. An assistant might directly perform a low-risk action, offer a suggestion without executing it, ask a clarifying question, continue observing, select a reversible alternative, or remain silent.

Importantly, these responses need not depend only on the predicted preference likelihood. Two actions with the same predicted likelihood may warrant different forms of initiative when one prediction is supported by extensive individual history and the other is inferred mainly from population-level patterns. A home assistant, for example, might automatically adjust lighting when it has repeatedly observed the user's preference in the same context, but merely suggest the adjustment when encountering that situation for the first time. In this way, uncertainty-aware initiative could help PSAs remain useful without treating every plausible prediction as sufficient justification for intervention.

\textbf{Questions that reflect what the assistant does not know.}
Active learning research shows that uncertainty can help systems identify observations that are particularly useful for personalization \cite{akbari2020personalizing,ardywibowo2019adaptive}. In future PSAs, this idea could influence not only whether the assistant asks a question, but also the content of that question.

Different knowledge gaps may naturally lead to different forms of interaction. When a prediction relies mainly on a population prior, the assistant might ask about a general preference: `Do you usually want the blinds closed when the afternoon sun reaches your desk?'' When the user's general preference is already known but the current situation is ambiguous, it might instead ask a contextual question: `Are you working on the screen right now?'' If the uncertainty comes from a temporarily unreliable sensor, waiting for another observation may be more appropriate than interrupting the user.

Such assistants could make preference elicitation feel less like repeated confirmation and more like a purposeful conversation. They might also allow users to choose how and when personalization occurs, for example by setting an asking budget, postponing questions, or identifying situations in which the assistant should not interrupt. The value of information would therefore be considered together with the cost of obtaining it from the user.

\begin{figure}
    \centering
    \includegraphics[width=\linewidth]{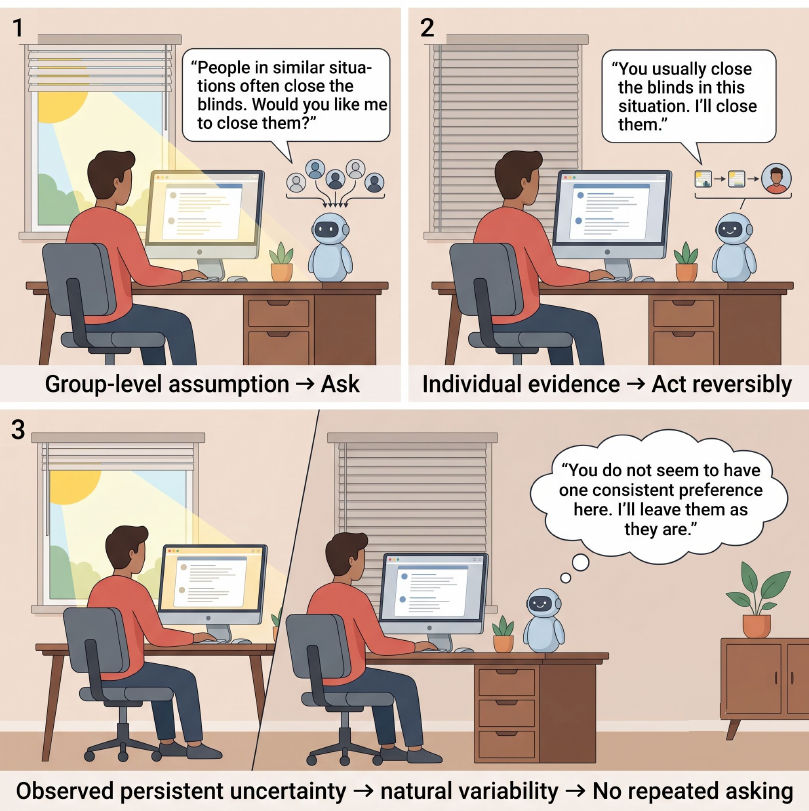}
    \caption{Future uncertainty-aware proactive assistants may respond differently to the same predicted preference likelihoods depending on the source of uncertainty.}
    \label{fig:placeholder}
\end{figure}

\textbf{Recognition that not every preference must become predictable.}
Human-motion prediction and uncertainty-aware learning distinguish uncertainty caused by insufficient knowledge from variability that is inherent in the observed behavior \cite{xu2021probabilistic,an2025disentangling}. A future PSA could similarly recognize that inconsistent choices do not always indicate that more personalization is needed.

A user may genuinely have no fixed preference between two actions, may consider several alternatives equally acceptable, or may decide based on factors unavailable to the assistant. In such cases, repeatedly asking questions or collecting additional data may not produce a more stable prediction. An uncertainty-aware assistant could instead recognize that the remaining ambiguity has persisted despite observation. It might stop trying to infer a single preferred action, rotate among acceptable options, choose the most reversible alternative, or ask only when the consequences are substantial.

\textit{Learning Progress Monitoring} \cite{hou2025beyond} similarly suggests that exploration should focus on uncertainty that is still being reduced rather than on prediction errors that remain irreducible. Applied to PSAs, this principle could help prevent personalization from becoming an endless effort to eliminate human variability. A future assistant might therefore know not only when it needs to learn more, but also when it has learned that no single stable preference exists.

\textbf{Explanations grounded in evidence provenance.}
Uncertainty attribution in trajectory prediction demonstrates that a system can identify which input sources contribute to a prediction \cite{itkina2023interpretable}. Future PSAs could use this capability to provide explanations that describe the basis of personalization rather than presenting a generic confidence score.
Instead of saying `I am 72\% confident,'' an assistant might explain, `I suggested this because you accepted it on the last three workdays,'' `I have not observed what you prefer when guests are present,'' or `This suggestion is currently based mostly on patterns from other users.'' These explanations reveal what the assistant believes it has learned and make incorrect assumptions easier to identify. They may also expose when a recommendation depends too heavily on group-level generalizations or on observations that the user does not consider relevant.

Such explanations could be paired with lightweight opportunities for correction, including `not true for me,'' `only when I am alone,'' or ``do not use this information.'' Uncertainty would then become an entry point for inspecting and repairing the assistant's model of the user. This may be more meaningful than displaying confidence alone, particularly because uncertainty displays and AI explanations do not automatically produce appropriate trust and may sometimes increase reliance on incorrect suggestions \cite{bhatt2021uncertainty,bondi2022role,buccinca2021trust}.

\textbf{Multi-user assistants that distinguish uncertainty from disagreement.}
Trajectory forecasting has introduced collaborative uncertainty to describe uncertainty arising from interactions among multiple people \cite{tang2021collaborative}. Shared PSAs face a related challenge. An assistant may be uncertain because it does not know one person's preference, because it cannot determine who is present, or because it knows that different users prefer incompatible outcomes. These situations may appear similar at the prediction level but call for substantially different responses.

Future multi-user PSAs could represent each affected person's preference separately from the procedure used to reconcile those preferences. When one person's preference is missing, the assistant might ask that person directly. When the users' preferences conflict, it might surface the disagreement, suggest a reversible compromise, or follow a previously agreed household rule. When the assistant is uncertain about who will be affected, it might avoid acting until the relevant users are identified.

This distinction could also help prevent shared assistants from treating the most frequently observed user as representative of the entire group. Existing work on collaborative assistants and interactive negotiation suggests that assistants can support group decisions rather than silently resolving them \cite{liu2024xiang,wang2026seesawbot,zhou2024bring}. An uncertainty-aware PSA could therefore serve not only as a predictor of collective preferences, but also as a mediator that makes missing representation and genuine disagreement visible.

\textbf{Uncertainty that evolves through long-term interaction.}
The uncertainty of a PSA is unlikely to remain fixed. It may decrease as the assistant observes recurring routines, increase when the user enters an unfamiliar situation, and change when preferences shift over time. A future uncertainty-aware assistant could therefore treat personalization as an ongoing interaction rather than a one-time learning phase.

Its behavior might become more autonomous in familiar, low-risk situations while remaining cautious in new or consequential contexts. It could recognize when an established pattern is no longer reliable and temporarily return to suggesting or asking rather than acting. It might also remember which questions have already been answered, avoiding repeated requests unless there is evidence that the user's preference has changed.

This longitudinal perspective broadens the evaluation of uncertainty aware PSAs beyond predictive calibration alone. The quality of such an assistant would also be reflected in the number and timing of its interruptions, the consequences of incorrect actions, the ease with which users can repair its assumptions, its improvement after feedback, and its treatment of changing or variable preferences. Users' understanding of when the assistant knows their preferences and when it is extrapolating would be similarly important. These concerns connect uncertainty estimation directly to established HCI questions surrounding agency, appropriateness, trust calibration, and user--system gaps \cite{oh2024better,kim2024navigating}.

Together, these possibilities envision uncertainty estimation as an interaction resource rather than solely a modeling technique. A future uncertainty-aware PSA would not merely attach confidence to its predictions. It would adapt its initiative, seek information selectively, recognize irreducible preference variability, reveal the evidence behind its assumptions, and distinguish missing knowledge from interpersonal disagreement. Such capabilities could support proactive assistance that is simultaneously more personalized, more understandable, and more respectful of the people whose lives it seeks to assist.

\bibliographystyle{ACM-Reference-Format}
\balance
\bibliography{references}


\end{document}